\def\year{2022}\relax
\documentclass[letterpaper]{article} % DO NOT CHANGE THIS
\usepackage{aaai22}  % DO NOT CHANGE THIS
\usepackage{times}  % DO NOT CHANGE THIS
\usepackage{helvet}  % DO NOT CHANGE THIS
\usepackage{courier}  % DO NOT CHANGE THIS
\usepackage[hyphens]{url}  % DO NOT CHANGE THIS
\usepackage{graphicx} % DO NOT CHANGE THIS
\usepackage{multirow}
\usepackage{multicol}

\usepackage{natbib}  % DO NOT CHANGE THIS AND DO NOT ADD ANY OPTIONS TO IT
\usepackage{caption} % DO NOT CHANGE THIS AND DO NOT ADD ANY OPTIONS TO IT
\DeclareCaptionStyle{ruled}{labelfont=normalfont,labelsep=colon,strut=off} % DO NOT CHANGE THIS
\usepackage{algorithm}
\usepackage{algorithmic}
\usepackage{todonotes} % DELETE THIS LINE FROM FINAL SUBMISSION
\usepackage{newfloat}
\usepackage{listings}
\usepackage{comment}
\floatstyle{ruled}
\newfloat{listing}{tb}{lst}{}
\floatname{listing}{Listing}

\newcommand{\eg}{{\textit{e.g.}}}

\title{Predictive Maintenance for General Aviation Using Convolutional Transformers}
\author{
    Hong Yang, 
    Aidan LaBella,
    Travis Desell
}
\affiliations{
    Golisano College of Computing and Information Sciences\\
    Rochester Institute of Technology\\
    20 Lomb Memorial Dr.\\
    Rochester, New York 14623\\
    \{hy3134,\ apl1341,\ tjdvse\}@rit.edu\\
}

\begin{document}

\maketitle

\begin{abstract}
Predictive maintenance systems have the potential to significantly reduce costs for maintaining aircraft fleets as well as provide improved safety by detecting maintenance issues before they come severe. However, the development of such systems has been limited due to a lack of publicly labeled multivariate time series (MTS) sensor data. MTS classification has advanced greatly over the past decade, but there is a lack of sufficiently challenging benchmarks for new methods. This work introduces the NGAFID Maintenance Classification (NGAFID-MC) dataset as a novel benchmark in terms of difficulty, number of samples, and sequence length. NGAFID-MC consists of over 7,500 labeled flights, representing over 11,500 hours of per second flight data recorder readings of 23 sensor parameters. Using this benchmark, we demonstrate that Recurrent Neural Network (RNN) methods are not well suited for capturing temporally distant relationships and propose a new architecture called Convolutional Multiheaded Self Attention (Conv-MHSA) that achieves greater classification performance at greater computational efficiency. We also demonstrate that image inspired augmentations of cutout, mixup, and cutmix, can be used to reduce overfitting and improve generalization in MTS classification.  Our best trained models have been incorporated back into the NGAFID to allow users to potentially detect flights that require maintenance as well as provide feedback to further expand and refine the NGAFID-MC dataset.
\end{abstract}

\section{Introduction}

In the domain of aviation, especially for small scale general aviation fleets, aircraft maintenance is performed with fixed schedules or after some maintenance issue is detected during operation of an aircraft. \emph{Predictive maintenance} techniques can be performed to reduce cost, improve machinery performance and life, as well as mitigate risk and increase safety. The majority of published literature covers non neural network methods ~\cite{PredModel-CARVALHO2019106024}.  Machine learning presents the potential to predict maintenance issues by measuring anomalies or degradation of multivariate time series (MTS) sensor data; however this has been limited by the proprietary nature of most flight data, with the further issue of acquiring the data necessary to label flight data with and without specific maintenance issues.

% While the abundance of multivariate temporal data has enabled significant advances in MTS analysis in a wide variety of fields, \eg, medical \cite{li2018tatc}, industrial \cite{christ2016distributed}), and energy \cite{cavraro2017power}, current literature lacks an evaluation of MTS methods for extremely long sequences (greater than 1024 time steps) from labeled datasets of significant size (more than 1000 datapoints). This paper utilizes data from the National General Aviation Flight Information Database (NGAFID) and the MaintNet project to create a new large scale labeled MTS benchmark, the NGAFID Maintenance Classification dataset (NGAFID-MC), consisting of over 7,500 labeled flight sensor data files\footnote{https://www.kaggle.com/hooong/ngafid-mc-20210917}, to enable the development of predictive maintenance systems for aviation. This dataset is used to investigate a new neural network architecture, a set of augmentation techniques for multivariate time series (MTS) classification, and provide a Google Colab Notebook for anyone to fully replicate the results of our experiments\footnote{Colab Notebook at tinyurl.com/b35mxv98}.Finally, the best models are used in NGAFID to classify flights and collect user feedback to further refine the models and expand the NGAFID-MC dataset.

While the abundance of multivariate temporal data has enabled significant advances in MTS analysis for a wide variety of fields, current literature lacks an evaluation of MTS methods for non synthetic, extremely long sequences (greater than 1024 time steps) from large labeled datasets (more than 5000 datapoints)~\cite{fawaz2019deep}. This paper utilizes data from the National General Aviation Flight Information Database (NGAFID) and the MaintNet project to create a new large scale labeled MTS benchmark, the NGAFID Maintenance Classification dataset (NGAFID-MC), with over 7,500 labeled flight sensor data files\footnote{https://www.kaggle.com/hooong/ngafid-mc-20210917}, for development of predictive maintenance systems for aviation.

Using this dataset, our results show that previous MTS classification methods face great difficulty in classifying pre and post maintenance flights. We also demonstrate that a new Convolutional Multiheaded Self Attention architecture can better capture complex temporally distant relationships within NGAFID-MC and leverage them for better classification performance and computational efficiency. We also demonstrate the need for robust augmentations and introduce a set of MTS augmentations that improve generalization. We provide a Google Colab Notebook for anyone to fully replicate the results of our experiments\footnote{https://tinyurl.com/b35mxv98}. Finally, our best trained models have been reincorporated into the NGAFID to inform users and collect their feedback to further refine the models and expand the NGAFID-MC dataset.

%MTS data is usually collected over predefined intervals (1 hour, 1 day, etc) at predefined frequencies (per minute, per second, etc) across a set of sensors (temperature, oxygen level, etc). 

%General aviation includes all non-military flights that are not operating regularly scheduled routes. These flights tend to be more dangerous than the commercial cargo and passenger flights \cite{aviationaccident}. Advances in technology have allowed for affordable recording of flight data, including data from various sensors throughout the aircraft. The National General Aviation Flight Information Database stores records of such flights. The ability to identify maintenance issues early based on flight data along can greatly improve pilot safety. 

\section{Related Work} 

Several methods have been developed for MTS classification, for a review see \cite{fawaz2019deep}. Notable non-deep learning methods include distance based k-nearest neighbors by \cite{orsenigo2010combining} and Dynamic Time Warping KNN by \cite{seto2015multivariate}. For deep learning methods, well performing MTS classifiers tend to utilize some combination of Recurrent Neural Network (RNN) and Convolutional Neural Network (CNN) methods, \eg~\cite{karim2017lstm}, or Temporal CNN (TCNN) methods  \cite{wang2017time}. However, RNN methods struggle with long sequences due to the vanishing gradient problem \cite{le2016quantifying}. TCNN methods perform well for MTS classification \cite{assaf2019mtex}, but may struggle when relevant features are temporally sparse and related.

Transformer models have been used in sequence tasks, such as NLP by \cite{devlin2018bert} and MTS prediction by \cite{zhou2021informer}. They do not suffer the vanishing gradient problem described by \cite{le2016quantifying}, allowing them to learn more temporally distant relationships. Application of transformer models and their underlying Multiheaded Self Attention (MHSA) mechanisms may lead to performance gains compared to RNNs. 

\cite{fawaz2019deep} notes that time series augmentation lacks a thorough study, compared to NLP and Computer Vision. Augmentations techniques in Computer Vision, such as cutmix \cite{yun2019cutmix}, may be applicable to MTS data. 

The datasets used by \cite{fawaz2019deep} do not exceed 1024 timesteps, except the WalkVsRun dataset consisting of only 28 training and 16 test examples. To the authors' knowledge, there are no MTS datasets that are not simulated, have greater length than 1024, and have labeled examples greater than 5000. Datasets meeting this criteria provide more realistic benchmarks for many real world applications, especially those related to engineering systems and predictive maintenance.

%\section{Maintenance Record Clustering}

%\section{Maintenance Need Classification}

\section{Dataset and Data Collection}

The NGAFID serves as a repository for general aviation flight data, with a web portal for viewing and tracking flight safety events for individual pilots as well as for fleets of aircraft~\cite{karboviak2018classifying}. The NGAFID currently contains over 900,000 hours of flight data generated by over 780,000 flights by 12 different types of aircraft, provided by 65 fleets and individual users, resulting in over 3.15 billion per second flight data records across 103 potential flight data recorder parameters. Five years of textual maintenance records from a fleet which provided data to the NGAFID have been clustered by maintenance issue type and then validated by domain experts for the MaintNet project~\cite{akhbardeh-etal-2020-maintnet}. Flights were extracted from the NGAFID and labeled as before or after the date of the maintenance action, creating a MTS dataset for training predictive maintenance models.

MaintNet's maintenance record logbook data was clustered into 39 different maintenance issue types. Because some issues occurred very rarely, this work focused on the two largest clusters, representing the most common maintenance issues: cluster 28 (C28): intake gasket leak/damage and cluster 37 (C37): rocker cover loose/leak/damage. The C28 and C37 clusters contain 1674 and 1248 maintenance records, respectively. Using the tail number from these maintenance records, the five flights preceding any of these maintenance records were exported from the NGAFID to represent flight data relating to those maintenance issues.  To provide a robust set of ``good'' flights without maintenance issues to compare these against, the five flights after the maintenance issues were exported as well, unless they were within 5 flights of any other maintenance issue. Flights shorter than 30 minutes were excluded as these are typically do not involve any actual flight. Flights were then further filtered within 2 days of maintenance (before and after). As the maintenance records only provided a day (and not a time) of action, flights occurring on the same day as maintenance were excluded as it was not possible to determine if they occurred before or after maintenance.

% This resulted in a benchmark dataset containing 7,505 flight data files representing 11,500 hours of Cessna 172S flight data, with each flight data file in this dataset consisting of data from 23 sensors (internal, external and operational sensors, \eg, engine RPM, oil temperature, oil pressure, gasket temperature, airspeed, pitch, roll, outside air temperature) recorded every second, with each flight labeled as pre or post maintenance.  Flights were additionally labeled by how many flights before or after the maintenance record they occurred. \hong{The number of flights before after is not used, only the number of days. The public data release doesn't have number of flights before after.} Flights are split for the two maintenance issues resulting in 1432 pre and 984 post examples for C37 and 2814 pre and 2275 post examples for C28.

This resulted in a benchmark dataset containing 7,505 flight data files representing 11,500 hours of Cessna 172S flight data, with each flight data file in this dataset consisting of data from 23 sensors (internal, external and operational sensors, \eg, engine RPM, oil temperature, oil pressure, gasket temperature, airspeed, pitch, roll, outside air temperature) recorded every second, with each flight labeled as pre or post maintenance. Flights are split for the two maintenance issues resulting in 1432 pre and 984 post examples for C37 and 2814 pre and 2275 post examples for C28.

\section{Background}

A major goal of this work is to be able to classify flights as problematic (leading to some maintenance issue), or in good condition (post maintenance). Three factors make this dataset challenging.  First is the sequence length, often exceeding 3600 time steps. Second is the nature of the prediction task, where the goal is to detect features relevant for classification. Third is the significant impact of unobservable variables, such as pilot actions, on the engine outputs.

To formalize the problem, we seek to predict the probability that a time series was generated by a pre or post maintenance flight given the flight sensor data. This can be expressed as \(P(Y_i | X_i)\). We have access to the variable \(X_{imt}\) as a matrix containing the flight sensor data, with \(imt\) representing the \(i\)th flight's \(m\)th variable at timestep \(t\). \(Y_i\) represents the \(i\)th flight's pre or post maintenance state as 1 and 0, respectively. \(U_{it}\) represents the pilot's actions for the \(i\)th flight's timestep \(t\). This unknown variable \(U\) is significant because it changes our understanding of the function generating \(X\) to \(f(U_i, Y_i) = X_i\). A pilot's actions can impact \(X_{imt}\) more than maintenance state of the aircraft.

We cannot construct a model to to predict \(X_{im(t+1)}\) using only the past timesteps of \(X_i\) due to the impact of \(U_i\). Similarly, a compressed representation \(c(X)\) may be useless for classification because it must first explain variance caused by \(U\). The authors believe that non-deep learning methods will struggle to perform well in these conditions.

This dataset provides an exciting challenge compared to industrial datasets, such as power plant data, because it measures a dynamic system that changes arbitrarily in a largely uncontrolled and inconsistent environment. Routine flight operations, such as landing and takeoff, can vary significantly from flight to flight due to the experience of the pilot, the weather, and wind conditions. We hope this dataset can serve as a challenging benchmark for MTS classification. 

\subsection{Model Architecture and Training}

\subsubsection{Augmentation}

To address the limitations of the size of this dataset, we looked into augmentations for MTS data. We considered only basic domain augmentation methods based on the taxonomy proposed by \cite{wen2020time}, as advanced domain augmentations are too complex, requiring one to train generative models. Basic time domain methods, as described by \cite{le2016data} and \cite{wen2020time}, include window slicing (training on slices of the MTS) and window warping (reducing or extending the length of a segment of the MTS). These methods were not seen to be applicable as window slicing should fail if features for classification are temporally distant and infrequent, and window warping may not be applicable if the data is not sinusoidal in nature, which this data is not.

The authors of this paper decided to explore new basic domain augmentations, inspired by highly effective augmentations for image classification. We consider the ideas cutout from \cite{devries2017improved}, mixup from \cite{zhang2017mixup}, and cutmix from \cite{yun2019cutmix}. To our knowledge, this paper is the first to evaluate the these augmentations on MTS, due to their absence in the surveys on MTS augmentation by \cite{wen2020time} and \cite{iwana2021empirical}.
 
Temporal cutout selects a random time segment and set of channels from a MTS and sets selected values to 0. Temporal cutmix selects a random time segment from the first MTS and a random time segment of a random second MTS of any label. It then selects a random set of channels and replaces the first MTS's segment's channel values with those of the second. Temporal mixup multiplies a randomly selected set of channels in the first MTS by a value, $m$, and then adds all values from the same set of channels from a randomly selected second MTS of any label multiplied by $1 - m$.

\subsubsection{Convolutional Multi-Headed Self Attention}

Multi-Headed Self Attention (MHSA) modules were popularized by \cite{devlin2018bert} for usage in Natural Language Processing and by \cite{dosovitskiy2020image} for Computer Vision, but published usage of MHSA in MTS data is less common than the previous two applications. Both \cite{song2018attend} and \cite{russwurm2020self}  use MHSA for MTS classification, but neither implements a 1D convolution followed by MHSA. \cite{karim2019multivariate} utilizes attention mechanisms in an Attention-LSTM network, but not MHSA. The benefits of a low level convolution prior to MHSA is demonstrated by \cite{gulati2020conformer}, where convolutions can capture basic local relationships with high efficiency while MHSA handles global relationships. 

At the time of writing, the authors are not aware of any publication that evaluates a convolutional MHSA model for MTS classification. This model implements attention layers that mimic the functionality of the encoder layers present in BERT \cite{devlin2018bert}. Instead of token embeddings, the model generates sequence embeddings with the use of 1D convolutions along the temporal dimension. These learnable sequence embeddings capture local relationships and to compress the MTS to a shorter length. 

MHSA modules offer significant benefits over LSTMs when applied to MTS. In particular, \cite{zhou2021informer} has shown the capacity for MHSA to model long term relationships in time series data. Furthermore, when applied to longer sequences, MHSA avoids the problems associated with a vanishing gradient as described by \cite{le2016quantifying}. Not only does it better model long term relationships, there is a significant computational efficiency over RNNs. 

\begin{figure}[t]
    \centering
    \includegraphics[width=0.47\textwidth]{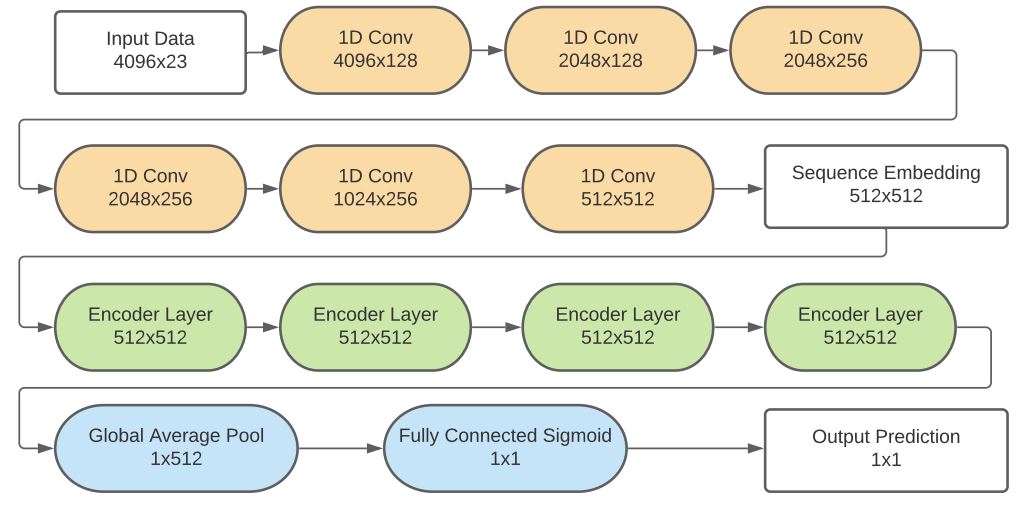}
    \caption{Layers and output shapes of the Conv-MHSA model. The first dimension represents time and the second represents channels. Note that white the boxes do not perform an operation, but mark significant states in the network.}
    \label{fig:arch}
\end{figure}

The Conv-MHSA model evaluated in this paper (see Figure~\ref{fig:arch}) uses a series of 1D convolutions to reduce the temporal resolution from 4096 to 512 and then employs 4 stacked  MHSA encoder layers with 8 heads each and 64 dense units per head. The output is globally average pooled and fed to a dense layer for classification.

\subsubsection{Convolutional Long Short Term Memory Networks}

\cite{keren2016convolutional} present a 1 dimensional convolutional LSTM as an enhancement to the traditional RNN. By using a 1 dimensional convolution, it is possible to extract features from the sequence before the LSTM layers and reduce MTS temporal resolution.

We consider two Conv-LSTM models. The first, referred to as Conv-LSTM, utilizes the same series of 1D convolutions as the Conv-MHSA model, but instead employs 4 stacked Bidirectional LSTMs with 512 units. The output is globally average pooled and fed to a dense layer for classification. The second Conv-LSTM is referred to as EX-Conv-LSTM, which utilizes 2 additional 1D convolutions before the stacked Bidirectional LSTMs to further reduce the temporal resolution to 128.

\subsubsection{Convolutional GRU Variational Auto Encoders}

Variational Autoencoders (VAE) were popularized by \cite{an2015variational} for anomaly detection. VAE's assume that a system's observable outputs \(X\) can be described via a vector or embedding \(E\), generated by the encoder component of the VAE model. When \(X\) cannot be described via a model generated embedding \(E\), it indicates anomalous activity. The ability to describe \(X\) via \(E\) is based on the ability of a decoder model to reconstruct \(X\) using only \(E\) and is measured as the reconstruction error. VAE learn their embeddings as a Gaussian distribution, with Kullback–Leibler divergence (KLD) for regularization. \cite{an2015variational} shows better performance for VAEs over standard Autoencoders.

GRU based VAE models have been employed by \cite{guo2018multidimensional} for anomaly detection in MTS data. For classification, this approach trains on within class data (post maintenance) with the expectation that out of class data (pre maintenance) will have greater reconstruction error. 

We implemented a VAE-Conv-GRU that uses 1D convolutions to reduce the temporal resolution to 256, followed by a bidirectional GRU (BD-GRU) with 256 units, then a 1D convolution to reduce the temporal resolution to 128, followed by another BD-GRU with 512 units. An embedding of 512 mixture Gaussian distributions and 8 mixtures per distribution is generated from the BD-GRU outputs, regularized via KLD. The decoder structure matches the encoder, with 1D Transposed Convolutions for the purpose of expanding the temporal resolution. Augmentations were not used.

\subsubsection{Training Setup}

All results reported were generated using a Google Colab instance with a v2-8 TPU. All models were trained for 30 epochs using a batch size of 32, with 5-fold cross validation. Steps per epoch are 250 for MHSA and LSTM models, 1000 for VAE models, and 500 for extended training LSTM models.  Flights are truncated to the last 4096 time steps and padded to be of the same size. To ensure the validation data is a good measure of generalized performance, the validation data is only composed of flights from tail numbers (unique identifier for planes) not present in the training data. Classification models used an Adam optimizer with a decaying learning rate starting at 1e-5 for MHSA and 2e-5 for LSTM and VAE models used an Adam optimizer with a decaying learning rate starting at 1e-4. Each augmentation (temporal cutout, cutmix, and mixup) was performed on a MTS with a 40\% chance. The time segment length for cutout and cutmix was selected uniformly at random between 64 and 512. Each channel had a 30\% chance of being selected for cutout and cutmix. For temporal mixup, $m$ was selected uniformly at random between 0.6 and 0.9, and it was applied to all time steps, with channels being selected with a 40\% chance. 
\subsection{Results}
 
\subsubsection{Computational Efficiency}

We observe significant computational advantages in the training of the Conv-MHSA compared to all other models. When using a TPU, the training step time (time to train on 1 batch) of Conv-MSHA is at least 4x faster than Extra-Conv-LSTM and at least 15x faster than Conv-LSTM. The results are summarized in Table \ref{Compute Time}.

\begin{table}[t!]
\centering
\begin{tabular}{ |c|c|c| } 
 \hline
 Model & Step Time in ms & Parameters\\
 \hline 
 C.MHSA & 50 & 7.9M\\
 \hline 
 C.LSTM & 800 & 24.7M\\
 \hline 
 EX C.LSTM & 220 & 28.4M\\
 \hline 
 VAE-Conv-GRU &  130 & 18.3M\\
 \hline
\end{tabular}
\caption{Approximate Training Step Time in Miliseconds}
\label{Compute Time}
\end{table}

Some of these advantages in step time could be caused by TPUs, which utilize matrix multiplication units (MXU's). Performance may differ on GPU systems.

%These MXU's are particularly well suited for MHSA, which is composed of a series of matrix multiplications. The process of back propagation through time may not be optimized or cannot be optimized further for a TPU. These major differences in computational efficiency can significantly reduce costs. It should be noted that increasing the batch size or switching to a vector based processing unit (GPU's) may reduce the performance disparity. 

\subsubsection{Classification Performance}

We evaluate each model's Area Under the Curve score for Precision-Recall (PR) and Receiver Operating Characteristic (ROC). These threshold independent metrics better measure generalized model performance than accuracy. Accuracy (ACC) is excluded from analysis because it depends on defining a threshold for predictions, which may be misleading due to class imbalance. Binary Cross Entropy loss in also considered as a metric to evaluate model overconfidence in wrong predictions. Results for VAE-Conv-GRU models are excluded from the table due to poor performance. See Table \ref{results}.

\begin{table}[t!]
\centering
\small
\begin{tabular}{|l|l|l|r|r|r|r|}
\hline
 &            Model Type & A &   Loss &    ROC &     PR  & ACC \\ \hline
 \multirow{10}{*}{C28} & \multirow{2}{*}{C.LSTM} &   Y &  0.630 &  0.701 &  0.654 & 0.653 \\ \cline{3-7}
  &      &   N &  0.617 &  0.742 &  0.697 & 0.685 \\ \cline{2-7}
  &     \multirow{2}{*}{C.LSTM+} &   Y &  0.623 &  0.730 &  0.644 &   0.691  \\ \cline{3-7}
  &     &   N &  0.613 &  0.757 &  0.711 &  0.694  \\ \cline{2-7}
  &       \multirow{2}{*}{C.MHSA} &   \bf{Y} &  \bf{0.528} &  \bf{0.826} &  \bf{0.802} & 0.744\\ \cline{3-7}
  &     &   N &  0.557 &  0.819 &  0.792  &\bf{0.751}\\ \cline{2-7}
  &    \multirow{2}{*}{EX C.LSTM} &   Y &  0.612 &  0.725 &  0.678 & 0.667\\ \cline{3-7}
  &    &   N &  0.614 &  0.755 &  0.713 & 0.694\\  \cline{2-7}
  &  \multirow{2}{*}{EX C.LSTM+} &   Y &  0.608 &  0.764 &  0.699 & 0.718\\ \cline{3-7}
  &    &   N &  0.612 &  0.785 &  0.737 & 0.733 \\ \hline
 \multirow{10}{*}{C37} & \multirow{2}{*}{C.LSTM} &   Y &  0.643 &  0.674 &  0.567 & 0.655\\ \cline{3-7}
 &      &   N &  0.679 &  0.553 &  0.489 & 0.596\\ \cline{2-7}
 &     \multirow{2}{*}{C.LSTM+} &   Y &  0.635 &  0.723 &  0.639 &  0.693\\ \cline{3-7}
 &     &   N &  0.644 &  0.711 &  0.618 & 0.683  \\ \cline{2-7}
 &       \multirow{2}{*}{C.MHSA} &   \bf{Y} &  \bf{0.601} &  \bf{0.775} &  \bf{0.711} & \bf{0.723}\\ \cline{3-7}
 &   &   N &  0.680 &  0.559 &  0.485 & 0.590 \\ \cline{2-7}
 &    \multirow{2}{*}{EX C.LSTM} &   Y &  0.632 &  0.708 &  0.620 & 0.677\\ \cline{3-7}
 &   &   N &  0.640 &  0.709 &  0.608 & 0.680 \\ \cline{2-7}
 &  \multirow{2}{*}{EX C.LSTM+} &   Y &  0.639 &  0.731 &  0.643 & 0.699 \\ \cline{3-7}
 &  &   N &  0.651 &  0.714 &  0.619 &  0.681\\ \hline
\end{tabular}

\caption{Mean of the best metrics for each configuration. LSTM + models are trained for 500 steps per epoch. C. stands for Conv. A stands for augmented.}
\label{results}

\end{table}

Results indicate that Conv-MHSA models consistently perform better than Conv-LSTM models by a wide margin. Even when Conv-LSTM models are given twice the number of training steps, they fail to reach the performance of MHSA models.

\subsubsection{Classification using VAE-Conv-GRU}

While the VAE-Conv-GRU model is capable of achieving a validation Root Mean Squared error of 0.0338, it cannot predict pre or post maintenance. With mean squared error as the reconstruction loss for comparing within class and out of class examples, the PR-AUC and ROC-AUC values never exceed 0.55.

\section{Discussion}

\subsubsection{Temporally Distant Attention}

\begin{figure}[t]
    \centering
    \includegraphics[width=0.47\textwidth]{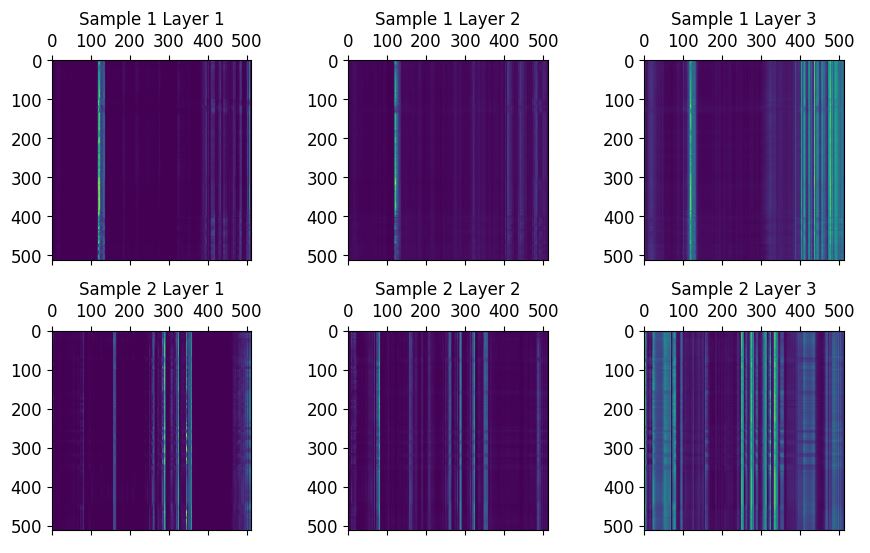}
    \caption{Attention maps illustrate how each time step attends to other timesteps in MHSA. This shows 3 MHSA Layers for 2 different datapoints from validation data. Y Axis represents Query and X axis represent Key. Sample 2 is positive and sample 1 is negative.  Bright sections are important timesteps that the model focuses on.}
    \label{fig:attentionmasks}
\end{figure}

To explore the question as to why MHSA can achieve better performance on this dataset compared to RNNs, it is important to observe how the various heads attend to different positions of the sequence. Figure \ref{fig:attentionmasks} is a visualisation of the 4 MHSA layers with multiple input datapoints. We can clearly observe in sample 0 that some layers are attending to time steps that are 300 units apart. An RNN model may have great difficulty in propagating information from time step 50 to time step 400 due to memory degradation and vanishing gradients. MHSA allows any time step to attend to any other time step and better capture temporally distant relationships.

To further show that the relationship between temporally distant features is necessary for classification, we attempted to train a Short-LSTM network using randomly sampled slices 128 time steps long. This network uses 4 stacked 512 unit bidirectional LSTMs followed by global average pooling and a dense sigmoid layer for classification. This Short-LSTM network did not perform significantly better than a fully random predictor, outputting random floating point values between 0 and 1. This demonstrates that random sub samples from the overall MTS is not sufficient.

\subsubsection{Augmentation}

The 3 augmentations of cutout, mixup, and cutmix, have similar functionality as dropout, described by \cite{srivastava2014dropout}. While it may seem counter intuitive to generate unrealistic sequences, these augmentations penalize the model for memorizing a small subset of time steps by removing or modifying them. Like their computer vision equivalents, these augmentations help models learn more resilient representations and improve generalization. 

\begin{figure}[t]
    \centering
    \includegraphics[width=0.47\textwidth]{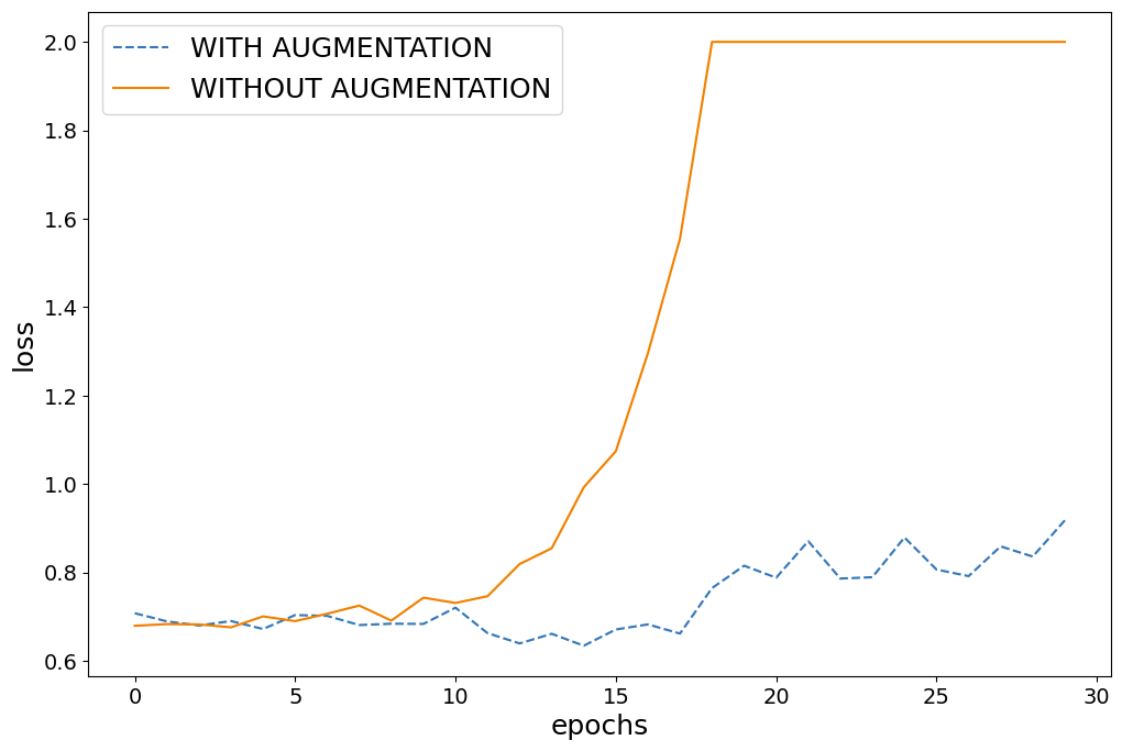}
    \caption{Validation Loss by epoch for Conv-MHSA model on the C37 dataset.}
    \label{fig:c37}
\end{figure}

%Figure \ref{fig:c37} shows how validation loss explodes for Conv-MHSA models on the C37 dataset without validation. 
Augmentations are particularly important for Conv-MHSA networks, which are prone to overfitting on small datasets. Conv-LSTM networks do not overfit and may not benefit from augmentation.  Results from experiments on Conv-MHSA models show a small advantage in the mean of all metrics when training on the C28 dataset, but a significant advantage when training on the C37 dataset. These differences are significant, such that Conv-MHSA models trained on C37 without augmentation perform not significantly better than random guessing. This is most likely caused by a difference in the dataset size, where C37 is about half the size of C28. This difference may also be caused by a difference in the nature of the data, where it is possible that C28 is easier to generalize on than C37. Figure \ref{fig:c37} shows the Conv-MHSA overfitting when training without augmentations on the C37 dataset. 

% All 3 augmentations operate on the same core principle of reducing reliance on specific segments of the time series, with cutout provided the most basic dropout like regularization. Mixup and cutmix also provides dropout like regularization, but they further also increase the diversity of the training set. By allowing differently labeled data points to be mixed together, the authors theorize that this further increases regularization by greatly penalizing the model for focusing a single segment of the time series. If a single segment is highly predictive, it can be cut from one label and placed in an opposite label forcing the network to reduce its reliance on such a segment. 

% With respect to Conv-LSTM models, we observe that augmentations hurt validation performance on the C28 dataset and improve validation performance on the C37 dataset. This is likely because the Conv-LSTM models encounter the underfitting problem on the C28 because it is sufficiently large and diverse. When Conv-LSTM models overfit, there is a significant advantage in using the augmentations.

\subsubsection{VAE and Reconstruction Loss}

\begin{figure}[t]
    \centering
    \includegraphics[width=0.47\textwidth]{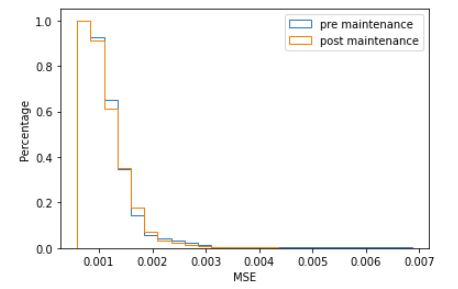}
    \caption{The Y axis indicates percentage of validation datapoints having more MSE than the number in X axis. The orange and blue lines represent pre and post maintenance, respectively. The distributions show no significant difference.}
    \label{fig:vae}
\end{figure}

Figure \ref{fig:vae} shows that the reconstruction loss is the same for both classes. This likely because a significant portion of the variance in \(X\) is caused by an unobservable variable \(U\). Any VAE model would first seek to learn how \(U\) impacts \(X\). Based on the analysis of MHSA on this dataset, there may be only a few, short segments of the MTS that are actually useful for classification. This suggests that VAE methods may struggle. 

%\travis{is it worth saying something along the lines that additionally the segments of the time series which would actually relate to the maintenance issue could be only short subsets of the entire time series and only appear within certain parameters, which could dramatically reduce any error signal from the VAE? or maybe that the VAEs are just capable of learning the appropriate patters pre and post maintenance and that's why there's low error?}

\begin{figure*}[t]
    \centering
    \includegraphics[width=0.6\textwidth]{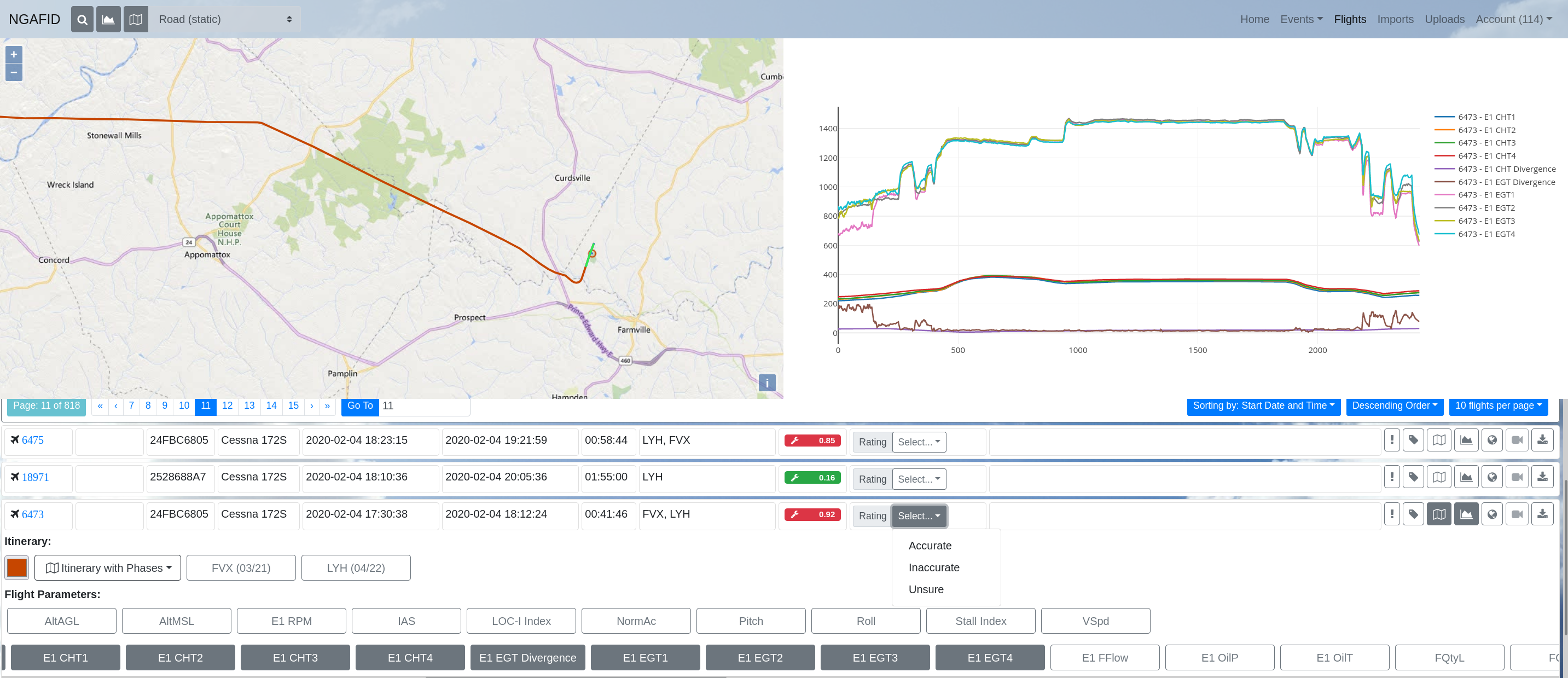}
    \hspace{5mm}
    \includegraphics[width=0.315\textwidth]{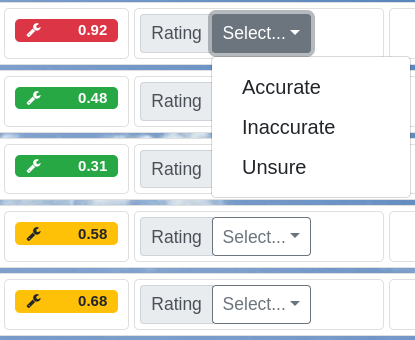}
    \caption{Screenshots showing the integration of the maintenance prediction models into the NGAFID user interface.}
    \label{fig:ngafid_screenshot}
\end{figure*}

\subsubsection{Future Research}

\noindent The NGAFID-MC dataset can be used to evaluate a wide variety of models and approaches, such as TCNN \cite{assaf2019mtex} and dynamic time warping \cite{seto2015multivariate}. Further studies can be performed on the full flight sequences, rather than only the last 4096 seconds of the flight. Future work should also evaluate mutliclass classification to identify which issue is present.  We also intend to expand the NGAFID-MC dataset with more maintenance issue cluster types, as well as refinements based on user annotations and as additional maintenance records are received.

The Conv-MHSA architecture performs much better than Conv-LSTM models on this dataset and it would be interesting to evaluate this on other datasets. It is also plausible that we can improve Conv-MHSA architecture by incorporating memory efficient methods described by \cite{kitaev2020reformer}. Additional work should be done on alternative loss functions for MTS classification, such as focal loss \cite{lin2017focal} and label smoothing \cite{szegedy2016rethinking}.

Cutout, mixup, and cutmix augmentations should be evaluated against other MTS augmentation methods and models. Due to the limited size of many MTS datasets and the cost to acquire data, further study into augmentation can increase the viability of MTS classification methods for general use. 

\subsubsection{Limitations}

There may be mislabeled datapoints due to the nature of airplane maintenance.  A reported maintenance issue may not be fully fixed or an issue was falsely identified by the pilot. If given resources, the authors would like to construct a small and rigorously annotated test set of data (1000 examples) with the help of domain experts. Additionally, the flights which occured on the day of maintenance were not included, these will be included in the future as they are annotated by domain experts.

\section{NGAFID Deployment and Integration}

The NGAFID provides a set of utilities for Flight Data Monitoring (FDM), which allow users to access the per-second time series data and perform various analytics. We added additional functionality to calculate and display the probability that a flight may require maintenance for the Cessna 172S aircraft type for flights exceeding 30 minutes (see Figure~\ref{fig:ngafid_screenshot}). This includes a feedback system was created to give users the ability to rate the accuracy of $P(Y_i|X_i)$ using a three-point scale (accurate, inaccurate or unsure), based on their knowledge of aviation and aircraft maintenance. This allows users to provide valuable feedback and labeled data for refining and improving future models. 

However, there are infrastructure challenges that need to be addressed before NGAFID can provide real time predictive maintenance alerts to improve safety and reduce costs. The main obstacle is the lack of wireless flight data transmission (WFDT), which is more common in commercial aviation settings. The current data import process for the NGAFID occurs weekly and requires ground crews to manually extract and upload the data. NGAFID partner fleets are in the process of deploying WFDT systems that will allow the NGAFID to perform real time predictive maintenance, as the WFDT systems can upload data immediately after an aircraft lands and returns to the hangar.

\section{Conclusion}

We demonstrate the challenging nature of the NGAFID-MC dataset and its value for assessing various MTS approaches. While some datasets exceed NGAFID-MC in terms of datapoints or sequence length, the authors are not aware of any dataset that has both greater datapoints and sequence length. The authors are also not aware of any other MTS dataset that tracks a dynamic system that changes arbitrarily in a largely uncontrolled and inconsistent environment. Furthermore, we demonstrate that this dataset contains temporally distant relationships that previous MTS classification methods struggle with. We hope that the difficulty of this dataset will inspire new and better methods for MTS classification. 

We also introduce a more computationally efficient and performant architecture, the Conv-MHSA. This architecture can better capture temporally distant relationships in long sequences and it does so with at much greater computational efficiency than RNN methods. We also show that cutmix, cutout, and mixup augmentations can significantly improve generalization.

The ability to differentiate between pre and post maintenance flights leads can provide a significant benefit to the domain of general aviation. Early detection of maintenance issues has the potential to reduce long term maintenance costs by catching issues before they cause more serious problems. By detecting the need for maintenance one or two days prior to maintenance, we can minimize the amount of flight hours that a pilot spends on compromised aircraft, leading to increased safety. We have already incorporated preliminary models for maintenance classification for NGAFID, which will allow us to gather feedback from users to further refine and improve the early maintenance issue detection system. We hope that these tools will lead to increased safety and reduced costs for general aviation.

% the style guide allows for small text for references
% \fontsize{9.0pt}{10.0pt} \selectfont
\bibliography{aaai22}

\end{document}